\documentclass[letterpaper, 10 pt, conference]{ieeeconf}  

\IEEEoverridecommandlockouts                              

\usepackage[table,xcdraw]{xcolor}
\usepackage{etoolbox} 
\usepackage[utf8]{inputenc} 
\usepackage[T1]{fontenc}    
\usepackage[english]{babel} 
\usepackage[protrusion=true,expansion=true]{microtype}
\usepackage{comment}
\usepackage{cancel}
\usepackage{csquotes}
\usepackage{listings}
\usepackage{nameref}
\usepackage{paralist}
\usepackage{xspace}
\usepackage{makeidx}
\usepackage{pdfpages}
\usepackage{footnote}

\usepackage{hyperref}
\hypersetup{
    colorlinks=true,
    linkcolor=black,
    citecolor=black,
    filecolor=cyan,
    urlcolor=blue
}

\usepackage{flushend}
\usepackage{balance}

\usepackage{paralist}
\let\labelindent\relax 
\usepackage{enumitem}
\setlist[itemize]{noitemsep,leftmargin=*,topsep=0in}
\setlist[enumerate]{noitemsep,leftmargin=*,topsep=0in}

\usepackage{amsmath,amssymb} 
\usepackage{amsfonts}       
\usepackage{mathtools}
\usepackage{array} 
\usepackage{nicefrac}       
\usepackage{breqn}          
\usepackage{textcomp}
\usepackage{bm} 
\usepackage{pifont}
\usepackage[
  separate-uncertainty = true,
  multi-part-units = repeat
]{siunitx}

\usepackage{graphicx}
\usepackage{adjustbox} 
\usepackage{epsfig}

\usepackage{float}
\usepackage[font=small]{subcaption}
\usepackage[font=small,labelfont=bf]{caption}
\usepackage{lscape}      
\usepackage{wrapfig}

\usepackage{tikz}  
\usepackage{makecell}
\usepackage{overpic}
\usepackage{algorithm}
\usepackage[noend]{algpseudocode}

\usepackage{tabularx}
\usepackage{multirow}
\usepackage{multicol}
\usepackage{booktabs}   
\usepackage{array} 

\usepackage{longtable}
\usepackage{threeparttable}

\makesavenoteenv{tabular}
\makesavenoteenv{table}

\usepackage{titlesec}
\titlespacing{\section}{0pt}{0.3\baselineskip}{0.25\baselineskip}
\titlespacing{\subsection}{0pt}{0.2\baselineskip}{0.15\baselineskip}
\titlespacing{\subsubsection}{0pt}{0.05\baselineskip}{0.03\baselineskip}

\renewcommand{\paragraph}[1]{\vspace{0.2em}\noindent\textit{#1} --}

\newcommand\mybar{\kern1pt\rule[-\dp\strutbox]{.8pt}{\baselineskip}\kern1pt}

\newcommand{\framework}{\textsc{SuFIA}\xspace}
\newcommand{\nvidia}{\textsc{Nvidia}\xspace}

\newcommand{\simName}{\textsc{Orbit}-Surgical\xspace}

\title{\LARGE \bf \framework: Language-Guided Augmented Dexterity \\ for Robotic Surgical Assistants}

\author{
Masoud Moghani$^{1}$, 
Lars Doorenbos$^{2}$,
William Chung-Ho Panitch$^{3}$, \\
Sean Huver$^{4}$,
Mahdi Azizian$^{4}$,
Ken Goldberg$^{3}$,
Animesh Garg$^{1,4,5}$
\thanks{$^{1}$University of Toronto, $^{2}$University of Bern, $^{3}$University of California, Berkeley, $^{4}$NVIDIA, $^{5}$Georgia Institute of Technology}%
\thanks{\href{mailto:moghani@cs.toronto.edu}{\texttt{moghani@cs.toronto.edu}}, \href{mailto:animesh.garg@gatech.edu}{\texttt{animesh.garg@gatech.edu}}%
}
}

\begin{document}

\maketitle
\thispagestyle{empty}
\pagestyle{empty}

\begin{abstract}
In this work, we present \framework, the first framework for natural language-guided augmented dexterity for robotic surgical assistants. \framework incorporates the strong reasoning capabilities of large language models (LLMs) with perception modules to implement high-level planning and low-level control of a robot for surgical sub-task execution. This enables a learning-free approach to surgical augmented dexterity without any in-context examples or motion primitives. \framework uses a human-in-the-loop paradigm by restoring control to the surgeon in the case of insufficient information, mitigating unexpected errors for mission-critical tasks. We evaluate \framework on four surgical sub-tasks in a simulation environment and two sub-tasks on a physical surgical robotic platform in the lab, demonstrating its ability to perform common surgical sub-tasks through supervised autonomous operation under challenging physical and workspace conditions.

\noindent Project website: \href{https://orbit-surgical.github.io/sufia}{orbit-surgical.github.io/sufia}
\end{abstract}

\section{Introduction}

Recently, one prominent trend in surgery has been the increasing adoption of robotic surgical assistants (RSAs) in operating rooms. These RSAs are often controlled via local or remote teleoperation through a console by a trained human surgeon using hand controllers or other input peripherals, thereby enabling the surgeon to perform tasks with enhanced precision, dexterity, and control during an operation~\cite{hwang2020applying}. The teleoperated surgical procedures often involve tedious, repetitive, or time-consuming sub-tasks. Augmented dexterity in surgery holds the potential to simplify the surgical workflow, reduce surgeon fatigue, and improve patient outcomes~\cite{sen2016automating, lin2023end}.

Learning-based approaches such as reinforcement and imitation learning learn policies to solve specific surgical sub-tasks~\cite{JMLR:v24:23-0207, xu2021surrol}. However, complex, long-horizon surgical sub-tasks are often computationally expensive, require extensive domain knowledge and reward engineering, and involve time-consuming dataset curation. Furthermore, the lack of generalizability limits the utility of learning-based models in safety-critical applications where unseen, in-domain variations are prevalent. As a result, most surgical robotic platforms still lack any level of autonomous capabilities~\cite{attanasio2021autonomy}.

In recent years, Large Language Models (LLMs) have received considerable attention for their ability to respond naturally to textual prompts and have been integrated into various domains, including the field of robotics and autonomous agents~\cite{wang2023survey}. Language and vision models have demonstrated considerable promise in long-horizon robot planning and control~\cite{driess2023palm, brohan2022rt, brohan2023rt}. While these efforts still require pre-trained skills and motion primitives, they have demonstrated the potential of unified many-modality models for addressing a variety of complex tasks involving improved generalization to novel objects and unseen tasks.

\begin{figure}
    \centering
    \includegraphics{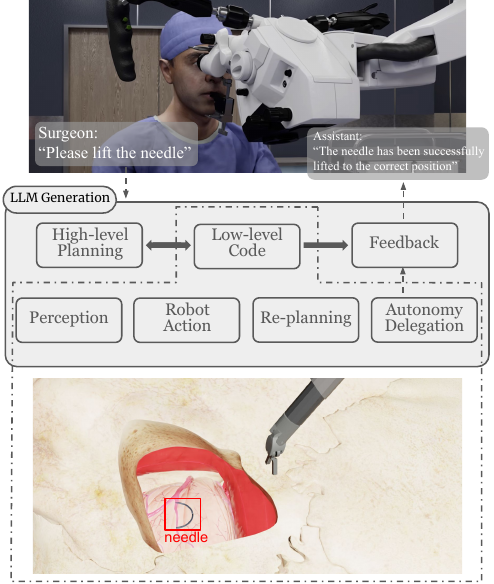}
    \caption{\textbf{An overview of \framework automating the lifting of a suture needle from a surgical site.} \framework receives commands from a surgeon in natural language and converts them to high-level planning and low-level control code. If a task requires object interaction, \framework queries a perception module for object state information and generates low-level trajectories and robot actions accordingly. \framework can assist a surgeon with open-ended tasks, such as moving the robot in a desired motion to help complete a surgical task. In times of inefficient information, \framework delegates full control back to the surgeon.}
    \label{fig:fig1}
\end{figure}

In surgical settings, LLMs have the additional potential to aid interaction between a human surgeon and a robot via natural language teleoperation. This empowers the surgeon with the ability to use both fine-grained manual control and autonomous natural language conversational control in commanding the RSA to perform a sub-task. This approach promises both more natural human-robot coordination and the potential for developing general-purpose models for autonomous surgery beyond the capability of current task-by-task automation approaches.

\begin{figure*}
    \centering
    \begin{minipage}[c]{0.69\textwidth}
        \includegraphics[width=\textwidth]{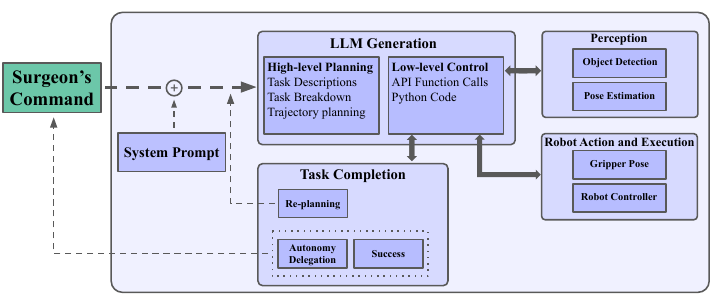}    
    \end{minipage}
    \,
    \begin{minipage}[c]{0.29\linewidth}
    \caption{\textbf{\framework architecture and workflows.} \framework enables a surgeon to naturally interact with the robot by either asking for a complete sub-task (e.g. ``pick up the needle and hand it over to the other arm'') or generating a trajectory to help with performing a task (e.g. ``move the needle 1 cm to the left''). \framework uses limited environmental knowledge in natural language (i.e. System Prompt) and scene understanding from a perception module to directly generate high-level plans and low-level sequences of gripper poses to interact with small-scale surgical objects. If \framework encounters difficulty in querying for an object or executing a necessary step to solve the surgical sub-task, it hands the control back to the surgeon for teleoperation.}
    \label{fig:fig2}
    \end{minipage}
\end{figure*}

In this work, we present \framework (Surgical First Interactive Autonomy Assistant), a framework for natural interaction between a human surgeon and a surgical robot to provide interactive surgical autonomy. As shown in Fig.~\ref{fig:fig1}, \framework takes in sub-task commands from a surgeon and outputs a high-level natural language task plan, as well as low-level Python code snippets for execution, if requested. A perception module grounds perceived surgical objects in the scene regardless of variations in shape, size, and pose and accounts for the characteristics of their often small, slender shapes. \framework also incorporates re-planning and human-in-the-loop control as safety measures. Our primary contributions are as follows:

\begin{itemize}
    \item A general formulation for natural language interaction between a surgeon and a robot.
    \item A language-based control approach to facilitate surgical sub-task implementations.
    \item A systematic evaluation of the generalization of our approach to various surgical sub-tasks, showing its performance and robustness for challenging workspace conditions.
\end{itemize}

\section{Related Work}

\subsection{Large Language Models for Robotics}
Large Language Models (LLMs) are state-of-the-art natural language processing systems built on the transformer architecture~\cite{vaswani2017attention}. LLMs are pre-trained with self-supervised objectives on vast amounts of text corpora, enabling these models to exhibit impressive language understanding and generation capabilities and perform a wide range of tasks. They are typically further fine-tuned with labeled data and RLHF to create general-purpose assistants~\cite{achiam2023gpt,liu2024visual} or more specialized models for use cases such as coding~\cite{luo2023wizardcoder,roziere2023code} or report generation~\cite{wang2023r2gengpt}.

In robotics, LLMs have been recently employed to address the high-level planning aspect of robotic control~\cite{liang2023code, vemprala2023chatgpt}. These models still require trajectory generators through cost or reward functions to compute the trajectory. Other works focused on leveraging LLMs to design reward functions~\cite{xie2023text2reward, ma2023eureka} to acquire complex skills via reinforcement learning. However, most of these research works perform well on predefined tasks and still require expensive training time to generalize. Building on a recent work~\cite{kwon2023language} that revealed the potential of LLMs to directly reason trajectory paths for robot arms, \framework incorporates LLMs to directly control the gripper poses to perform surgical sub-tasks. This enables the surgeon to naturally interact with the robot by asking for a complete task (e.g., pick the needle, insert the soft tube) or an open-ended task (e.g., move the needle in semi-circular motion) to help complete a sub-task. Our work differs from~\cite{kwon2023language} in that we do not rely on a separate object detector for validation, incorporate further safety mechanisms by delegation, and show results for surgical scenes, where we additionally study domain-relevant axes such as variations in needle shape.

\subsection{Surgical Augmented Dexterity}

Augmented dexterity has been attempted for several sub-tasks with varying levels of autonomy~\cite{nagy2020autonomous,ficuciello2019autonomy,attanasio2021autonomy} such as dexterous needle picking and handling~\cite{wilcox2022learning, lin2023end}, suturing~\cite{sen2016automating,krishnan2017transition}, and tissue manipulation~\cite{shin2019autonomous, nguyen2019manipulating, li2020super}. In particular, in contrast to full automation, an often-explored paradigm in surgical robotics is augmented dexterity~\cite{kyg}, in which minimal surgical sub-tasks are automated under human supervision, enabling more precise actuation with less effort expended.

However, these works largely rely on access to expensive, task-specific surgical hardware and software. In order to enable wider exploration of robotic automation in surgery, prior work has often focused on reducing the hardware barrier by adapting traditional robotic arm geometries for medical sub-tasks~\cite{shademan2016supervised, saeidi2022autonomous}, or designing novel, lower-cost, multi-purpose medical robotic systems~\cite{hannaford2012raven, kazanzides2014open, ranzani2015bioinspired}. Additionally, learning robust perception and control models for surgical tasks often requires gathering very large and expensive in-vivo datasets to avoid safety-critical failure cases~\cite{ozyoruk2021endoslam, 9153443}. In this work, we propose an alternate approach to this software barrier by relying on a general-purpose, natural language-guided framework for surgical augmented dexterity across multiple tasks.

\section{Problem Formulation}

We focus on a novel approach to surgical augmented dexterity. In contrast to previous methods, we are investigating the potential of a generalist framework using large language models to address surgical augmented dexterity rather than training individual models for isolated tasks. We now briefly detail the assumptions with respect to the environment and available tools in our work. We do not provide any policies, trajectory optimizers, or in-context examples to the LLM~\cite{kwon2023language}. Instead, we expect the LLM to reason over automating a benchmark simulated surgical sub-task with their internal knowledge and access to limited environment information through pre-defined function calls available in an API. All of our experiments are carried out on the da Vinci Research Kit (dVRK) robot platform~\cite{kazanzides2014open}. The initial position and orientation of the dVRK grippers are available from the robot controller.

We assume access to a single RGB-D camera with a known intrinsic matrix, allowing for transformation between the camera's perspective and the world coordinate space. With this, we design a perception module for the LLM to interact with and query object information. This module identifies and retrieves the pose information of objects present in the scene. For this, in simulation, we assume access to an instance segmentation model that, given an object name, outputs the segmentation maps of all instances of the queried object. In the physical experiments, we train a segmentation network based on the architecture from~\cite{dharmarajan2023robot} for a needle segmentation model.

\section{\framework}

We propose \framework, a framework for natural interaction between surgeons and robots. \framework uses a human-in-the-loop approach, allowing either complete sub-task autonomy or assistance in open-ended tasks to help surgeons achieve their desired goals. The architectural framework, workflow, and primary elements of \framework are shown in Fig.~\ref{fig:fig2}. The following sections elaborate on the specifics of \framework.

\subsection{LLM Generation and Planning}

Crucial to the effectiveness of any LLM-based system is the design of the prompt, as only changing the prompt format can already lead to large differences in performance~\cite{zhao2021calibrate}. We build upon~\cite{kwon2023language} who developed a single task agnostic prompt for performing low-level robot control for object grasping. We adapt it for surgical augmented dexterity with a four-part prompt, which consists of a role description and three core parts: the first part contains the API library available to the LLM; the second provides limited environment information (e.g. the state of the robot(s) to control and the orientation of the coordinate system); the third provides general instructions on how the LLM should generate the code, including the format of the desired output; the fourth describes prompt optimizations such as doing step-by-step reasoning~\cite{kojima2022large}.

\subsection{API Library}

The LLM has access to a library of functions that are available through an API. This API is documented in the main prompt, where for each function, its signature is given along with a brief description of its functionality~\cite{kwon2023language,singh2023progprompt}. The API library mainly manages interaction with the robot control and perception modules. The modular approach with the API provides \framework the flexibility to adapt its respective modules independently, enabling integration into new embodiments and environments, such as switching from simulation to physical experiments.

Specifically, the API library includes robot control functionalities to execute a trajectory, rotate or open/close the gripper of the specified robot arm, and return the control back to the surgeon. Furthermore, perception functions detect the world poses of objects within the environment and can validate whether an object is at the expected position.

\subsection{Perception}

While LLMs lack the capability to ground physical worlds~\cite{liu2022mind}, they can still reason over the required steps to interact with objects and plan for task execution. To do so, we design a perception module that enables the processing of observations of the environment obtained from a single RGB-D camera to provide the object states to the LLM generator. This workflow is enabled by the API function \emph{detect_object}, through which the LLM queries and interacts with the perception module to retrieve object information.

\emph{detect_object} takes as input the name of the object to detect. After obtaining a segmentation of the named object and projecting it to world coordinates with the camera intrinsic matrix, we compute the 3D bounding cube and obtain the location and orientation. Moreover, for circular objects such as needles, we fit RANSAC to provide the object parameters and compute candidates for the location and orientation of a suitable interaction point.

\subsection{Safety}

A critical issue in surgical robotics is the reliability and safety of the robot control. To this end, we implement two components tailored to improve these aspects:

\begin{enumerate}
    \item \textbf{Re-planning.} The original plan could become inappropriate due to, for example, mistakes in the planning or unforeseen circumstances, such as the gripper losing grip on the needle, in which case a new plan has to be devised. We encourage \framework to repeatedly use the \emph{verify_object} function to check whether the observed position of a given object matches the position expected by the framework. If the object being manipulated is not in the expected place, \framework re-plans the steps to complete the desired task given the updated knowledge of the environment.
    \item \textbf{Human-in-the-loop approach.} In some cases, the perception module cannot find the desired object. In this case, rather than continuing blindly, \framework proceeds by handing control back to the surgeon for teleoperation with the API function \emph{transfer_control}. \framework is also instructed to call this function when it does not know how to solve a certain (sub-)task rather than operating on insufficient information. Fig.~\ref{fig:fig6} illustrates an instance when the system is unable to execute a command properly and returns the control to the surgeon to adjust the environment or provide further instruction.
\end{enumerate}

Together, these two components enhance the safety and reliability of the assistant, which is crucial in the domain of surgical robotics. Note that the surgeon can also directly instruct \framework to re-take control of the robot, ensuring a smooth interplay between surgeon and robot.

\section{Experimental Results}

To empirically measure the efficacy of \framework, we perform experiments both in \simName, a high-fidelity surgical simulation framework, and on a dVRK platform in the lab.

\subsection{Experimental Setup}

We conduct our simulation experiments in \simName~\cite{yu2024orbit}, which accurately imitates joint articulation and low-level controllers of the real dVRK platform, supports contact-rich physical interactions between rigid and deformable objects, and provides high-fidelity rendering. Furthermore, \simName provides an interface for teleoperation, which enables the user to work together with \framework to solve a sub-task if needed. We use a camera sensor in \nvidia Omniverse to acquire $512 \times 512$ rendered RGB-D images and ground-truth semantic segmentation masks. In section~\ref{sec:langsam}, we will discuss the utility of general segmentation models to adapt the workflow to objects with various configuration in simulation.

Physical experiments are performed on a da Vinci Research Kit (dVRK)~\cite{kazanzides2014open} robot surgical assistant, using an Allied Vision Prosilica GC 1290 stereo camera pair for visual input. These cameras are capable of producing paired stereo frames at a resolution of 1280 $\times$ 960 at 33 fps. Real-world depth images are then subsequently obtained by passing image pairs through RAFT-Stereo RVC~\cite{jiang2022improved}, a state-of-the-art network for predicting image correspondences using optical flow, and then using the camera's calculated intrinsic matrix to retrieve depth from these point discrepancies. We find that this approach provides better empirical results than traditional depth cameras in our use case due to the small, reflective objects and short focal lengths involved in the surgical setting. To emulate the real-world conditions encountered in a surgical setting, our workspace consists of a 3-D Med suturing tissue phantom on a red background. The phantom is then wrapped in blue cloth to imitate the use of a surgical cover during operation. These physical experiments introduce additional challenges, including a more challenging perception task, estimation and control noise, and more complex physics.

Throughout this section, we use GPT-4 Turbo~\cite{achiam2023gpt} unless stated otherwise.

\subsection{Tasks and Evaluation Metrics}

\begin{figure}
    \centering
    \includegraphics{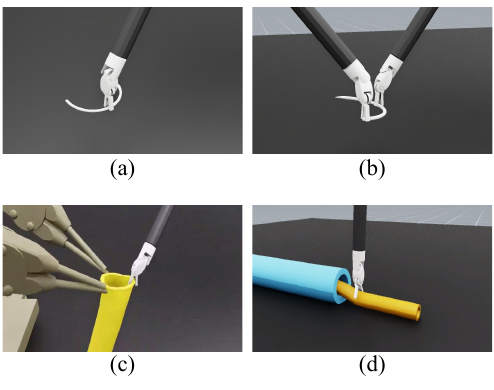}
    \caption{\textbf{Surgical sub-tasks.} (a) \texttt{Needle Lift}: lift a suture needle to a desired height, (b) \texttt{Needle handover}: pick and handover a suture needle, (c) \texttt{Vessel Dilation}: grip the vessel rim and dilate by pulling, (d) \texttt{Shunt Insertion}: insert a soft tube into larger vessel phantom. Best viewed in color.}
    \label{fig:fig3}
\end{figure}

\begin{figure*}
    \centering
    \includegraphics[width=\textwidth]{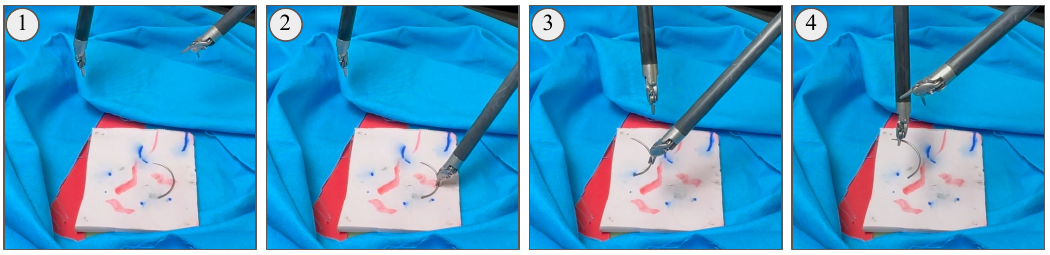}
    \caption{\textbf{Physical \texttt{Needle Handover} task}. (1) Starting workspace configuration. The needle is placed in a fixed position within the workspace, and the gripper positions are randomized. In this stage, the \framework LLM planner queries for and identifies the pose of the suture needle, determines which robot arm is closest to it, and plans a trajectory for that robot arm to reach the suture needle. (2) The closest robot arm approaches and grasps the suture needle. (3) The suture needle is lifted to a neutral handover position. At this stage, the \framework LLM planner detects the pose of the suture needle at the handover position and plans a trajectory for the second robot arm to approach the needle. (4) The second robot arm descends and grasps the needle, then the first robot arm releases the needle after the second robot arm has grasped it. We provide task videos at \href{https://orbit-surgical.github.io/sufia}{orbit-surgical.github.io/sufia}}
    \label{fig:fig4}
\end{figure*}

We demonstrate the generalizability of \framework by evaluating it across four distinct simulated surgical sub-tasks derived from \simName, as shown in Fig.~\ref{fig:fig3}. We additionally select two of the subtasks (Needle Lift and Needle Handover) for evaluation using our physical setup. Each sub-task poses unique challenges to show the robustness of the proposed workflow, described as follows:

\texttt{Needle Lift} -- In this task, the needle (N1 in Fig.~\ref{fig:fig5}) is initialized at a random position and orientation within the reach area from a single dVRK arm. The task is successful if the robot grasps and lifts the needle to a specified height above the table.

\texttt{Needle Handover} -- This task involves transferring a needle using a dual-arm dVRK setup. The needle is initially positioned randomly. The arm closest to the needle first grasps and lifts it to a specified handover location. Subsequently, the second arm reaches for the needle, grasps it, and takes it to a desired position. The task is successful if the needle is effectively transferred from the initial to the second arm.

\texttt{Vessel Dilation} -- In this task, a spring clamp assembly holds a soft vessel phantom from two points. The dVRK arm is required to grip the vessel rim from a third point facing the robot and dilate the vessel by pulling backward. A successful trial is defined if the robot fully dilates the vessel.

\texttt{Shunt Insertion} -- This task requires using a dVRK gripper to insert a shunt into a vessel phantom. The arm grasps the shunt from the middle, lifts it slightly, and then inserts it into a vessel phantom. The task is considered successful if, upon release by the grippers, the shunt remains inside the vessel phantom.

\subsection{\framework Evaluation}

We now discuss the effectiveness of \framework on solving the proposed surgical sub-tasks. \framework utilizes a perception module to localize the objects and proposes a sequence of sub-trajectories to perform the required task. We present the success rate for each sub-task for 10 trials in TABLE~\ref{tab:evaluation}. Overall, \framework is able to solve all proposed surgical sub-tasks requiring precise grasping of small surgical objects in simulation. Each task poses a unique challenge for automation, including object-gripper alignment and executing many steps to achieve successful results. In the \texttt{Vessel Dilation} task, all planning failures were due to not rotating the grippers to grasp the vessel's rim correctly. In the \texttt{Shunt Insertion} task, the planning failures were from incorrect lift height calculations before insertion.

We observed that the performance of \framework was relatively robust to the more complex physics and observation spaces of the physical environment, with 0 and 2 planning failures encountered during the \texttt{Needle Lift} and \texttt{Needle Handover} experiments, respectively. This aligns closely with the framework's performance in simulation. However, we found that hysteresis and encoder mismatch within the cable-driven dVRK resulted in variation between the commanded and actual gripper positions. Although \framework was often able to recover from the failures induced by this mismatch through its re-planning behavior, the lack of explicit servoing can result in dropping the needle during more complicated handovers.

\begin{table}[]
\resizebox{\columnwidth}{!}{
\centering
\setlength\extrarowheight{3pt}
\begin{tabular}{lcccc}
\toprule
\rowcolor[HTML]{CBCEFB} 
\cellcolor[HTML]{CBCEFB} & \cellcolor[HTML]{CBCEFB} & \cellcolor[HTML]{CBCEFB} & \multicolumn{2}{c}{\cellcolor[HTML]{CBCEFB}Failure Modes} \\
\rowcolor[HTML]{CBCEFB} 
\multirow{-2}{*}{\cellcolor[HTML]{CBCEFB}Experiment} & \multirow{-2}{*}{\cellcolor[HTML]{CBCEFB}Success Rate} & \multirow{-2}{*}{\cellcolor[HTML]{CBCEFB}Planning Steps} & (P) & (E) \\
\midrule
\midrule
\textbf{Sim Experiments} & & & & \\
\midrule
Needle Lift & 100 \% & 6 & 0 & 0 \\
\rowcolor[HTML]{EFEFEF} 
Needle Handover & 90 \% & 14 - 16 & 1 & 0 \\
Vessel Dilation & 60 \% & 6 - 8 & 3 & 1 \\
\rowcolor[HTML]{EFEFEF} 
Shunt Insertion & 70 \% & 8 - 9 & 3 & 0 \\
\midrule
\midrule
\textbf{Physical Experiments} & & & & \\
\midrule
Needle Lift & 100 \% & 6 & 0 & 0 \\
\rowcolor[HTML]{EFEFEF} 
Needle Handover & 50 \% & 14 - 18 & 2 & 3 \\
\bottomrule
\end{tabular}
}
\caption{\textbf{Evaluation} Success rate and planning steps required for surgical sub-tasks automation (10 trials for each experiment). Failure modes: (P) denotes planning and (E) denotes execution failures. Sim experiments are carried out in \textsc{Orbit}-Surgical, a high-fidelity surgical simulation framework. Physical Experiments are performed on a dVRK surgical platform.}
\label{tab:evaluation}
\end{table}

\subsection{Task Prompt Analysis}

Simple tasks such as \texttt{Needle Lift} require a simple prompt to function properly. The surgeon can specify a position to transfer the needle to or allow the LLM to determine a specific lift height above the table.

More sophisticated prompts are needed for tasks that require several steps for successful completion. In the \texttt{Needle Handover} task, the surgeon can provide additional notes for \framework to consider (e.g., "please note that for a handover, each robot should grasp the needle from the side closest to it."). The sequence in which the robot arms grasp and hand over to each other, as well as the location of the handover, can either be specified directly or left for the \framework to decide based on the distance to the needle or other environmental states.

The \framework planner may suggest unnecessary steps that may not be required for task completion and may potentially elongate task execution time. For instance, in the \texttt{Vessel Dilation} task, the vanilla prompt for dilating a vessel can sometimes lead to an additional step of "Lift the vessel slightly by moving the end-effector upwards to provide clearance from the table." The surgeon can provide additional information about the fact that the clamps are holding the vessel vertically to eliminate the suggestion of lifting steps in dilating the vessel. Similarly, in the \texttt{Shunt Insertion} task, additional information such as "please lift the small tube by a specific amount off of the table and horizontally insert it" helps to achieve better planning and execution.

Vision language models (VLMs) can also be incorporated in \framework to enhance the general visual understanding of the LLM planner. For instance, in the \texttt{Vessel Dilation} task, GPT4-Vision~\cite{achiam2023gpt} can provide the planner with environmental context regarding the orientation of the vessel phantom. In this example, the VLM response can complement the user prompt: \texttt{I see a vertical yellow tube on the right side of the image. It appears to be standing upright on one of its ends on a flat surface.} While useful for providing general visual context, similar to~\cite{skreta2024replan}, we find GPT4-Vision unreliable as a standalone perception module for detecting (small) objects' spatial states and omit it for the remainder of our experiments.

The prompts used for the tasks are as follows:

\paragraph{\texttt{Needle Lift}} "Pick up the needle and lift it."

\paragraph{\texttt{Needle Handover}} "Pick up the needle with the arm closest to it, move it directly to the handover location between the two arms, and keep holding the needle. Grasp the right side of the needle with the other robot arm, then right after that, release the needle from the first robot and stay put."

\paragraph{\texttt{Vessel Dilation}} "Grasp the vessel from its leftmost side with robot 0 and pull it backward to the left by 5 millimeters while holding on to it to dilate. When grasping the vessel, grasp it 15 millimeters below the left point."

\paragraph{\texttt{Shunt Insertion}} "Lift the small shunt from the middle and insert it into the left opening of the large tube. Approach the large tube from the left. Only lift the tube by 8 millimeters and move horizontally to insert."

\begin{figure}
    \centering
    \includegraphics[width=0.85\linewidth]{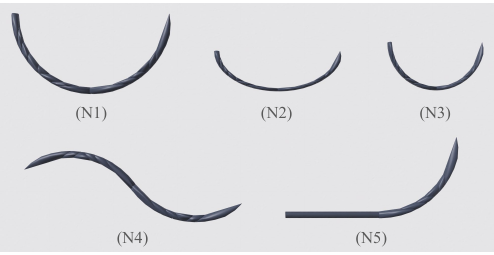}
    \caption{\textbf{Needle variations in simulation.} We consider five instances of simulated suture needles (N1 - N5) with various sizes and shapes to conduct the generalizability experiment in \simName.}
    \label{fig:fig5}
\end{figure}

\subsection{Perception Adaptation and Domain Variation}

\label{sec:langsam}
While we envision domain-specific perception models for applications of our framework in a given surgical environment (e.g., a needle segmentation network in our real-world physical experimentation), here we investigate whether a general-purpose segmentation model, LangSAM~\cite{langsam}, can be utilized by the perception module in simulated surgical environments to enhance the generalizability of the \framework framework to various object shapes.

We found that due to the slender shape of dVRK arms, the use of LangSAM in the perception module sometimes returned both the dVRK arm and the suture needle when prompted to find the "needle." However, descriptive adjectives (e.g. "round" or "small white needle") enable LangSAM to correctly identify and segment the suture needle well enough for \framework to generate the required steps to grasp and lift it. GPT-4 was also able to reason over the sizes of the segmented objects and properly determine the object of interest; here is an example of the LLM planner: "\texttt{there are two objects detected as "white needle," but only one of them has dimensions that match a needle (Width: 0.011, Length: 0.032, Height: 0.002). The second object's dimensions are too large to be the needle we are interested in. Therefore, we will focus on the first object with the correct dimensions.}"

To study the generalizability of the perception module across various needle sizes and shapes, we conduct a study with five different needles, three needles in different sizes (N1 - N3; Fig.~\ref{fig:fig5}) and two irregular shapes (N4 and N5; Fig.~\ref{fig:fig5}). TABLE~\ref{tab:evaluation-needles} shows the performance of \framework to lift various needles, which is robust to their shape and size.

\begin{figure*}
    \centering
    \includegraphics{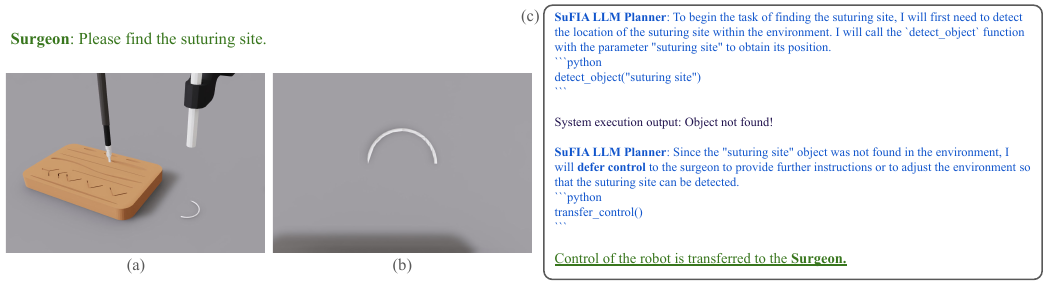}
    \caption{\textbf{Interactive human-in-the-loop approach.} (a) An overview of the environment showing the dVRK robotic arm and endoscope camera as well as a needle and a suturing pad in \simName, (b) RGB image from the endoscope camera focused on the needle as input to the perception module, (c) delegation of control back to the surgeon due to the inability of the system to identify a suturing site. The uncertainty and occlusion in a surgical scene might lead to undesired object localization and retrieval. A human-in-the-loop approach enables a fail-safe paradigm for interactive robotic surgical assistants.}
    \label{fig:fig6}
\end{figure*}

\begin{table}[]
\centering
\resizebox{\columnwidth}{!}{%
\begin{tabular}{lccccc}
\toprule
\rowcolor[HTML]{CBCEFB} 
Perception Module & N1 & N2 & N3 & N4 & N5 \\
\midrule
\midrule
Isaac Sim Camera & 5 / 5 & 4 / 5 & 5 / 5 & 5 / 5 & 4 / 5 \\
\rowcolor[HTML]{EFEFEF} 
LangSAM & 4 / 5 & 5 / 5 & 4 / 5 & 3 / 5 & 3 / 5 \\
\bottomrule
\end{tabular}%
}
\caption{\textbf{Domain variation evaluation in simulation.} We report the success rates for lifting suture needles with varied sizes and irregular shapes (suture needles N1 - N5) over 5 trial runs with two variations of the perception module.}
\label{tab:evaluation-needles}
\end{table}

\subsection{LLMs Investigation}

Here, we investigate the effect of different LLMs on the performance of the needle lift task. We use the same prompt and needle locations for all LLMs. As the error handling in \framework would, in principle, allow an LLM to keep trying endlessly until it generates code where no exceptions are raised, we limit the number of errors to five before terminating the program.

None of the open-source LLMs can perform the simple task of needle picking and have a hard time following the instructions in the prompt. All models struggle with understanding that \texttt{detect_object()} will print its result rather than return it as a variable in a Python script. When faced with errors, Mixtral~\cite{jiang2024mixtral} typically only outputs updated code snippets when asked to improve a code block rather than the whole code. CodeLlama~\cite{roziere2023code} calls many undefined functions, such as \texttt{get_end_effector_pose()}, despite the end-effector pose being given in the prompt. Llama 2~\cite{touvron2023llama} has a variety of mistakes related to understanding the steps in the task, such as forgetting to close the gripper or moving it down before lifting the needle.

GPT3.5 Turbo similarly misunderstands \texttt{detect_object()}, often assigning its value to a variable called \texttt{needle_position}, despite the prompt stating the function does not return anything. Beyond that, GPT3.5 Turbo does consistently define a proper plan to lift the needle, but even when it calls \texttt{detect_object()} correctly, the information is not incorporated successfully.

All in all, in our experiments, only GPT-4 Turbo could follow all instructions and appropriately plan and execute the relatively simple task of lifting a suture needle.

\subsection{Re-planning}

\begin{figure}
    \centering
    \includegraphics{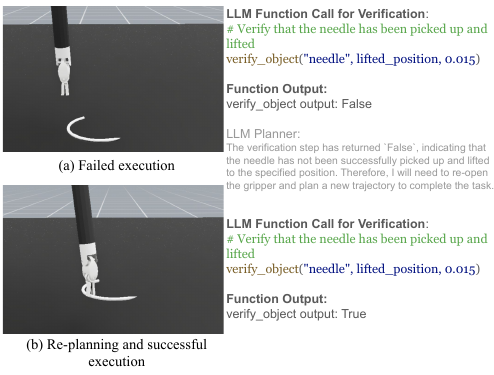}
    \caption{\textbf{Re-planning.} (a) A failed execution as a result of not finding a proper grasping point on the needle. The LLM verification step indicates that the task was not executed successfully. (b) LLM planner proposes a new plan to re-identify and lift the needle by the robot.}
    \label{fig:fig7}
\end{figure}

To illustrate the benefits of our safety modules, we provide an example in the \texttt{Needle Lift} environment in Fig.~\ref{fig:fig7}. In the first row, \framework executes the plan it came up with to perform the task desired by the user, i.e., orienting its gripper with the needle, moving to a position where it can grab it, and picking it up. While picking it up, we move the needle to a different position. Because \framework validates the expected and observed position of the objects it manipulates, it correctly identifies the needle is not where it should be. Based on the newly observed state, \framework devises a new plan to proceed with the user instruction, finally lifting it to the desired height.

\subsection{Limitations}

The current best results are obtained with API calls to GPT-4 Turbo. Due to the generation speed of OpenAI's API, \framework does not operate in real-time; for real-world needle handover trials, the API calls invoked during planning (including sensing and replanning) took an average of 25.5 seconds to complete, out of an average total trial time of 61.4~seconds. However, with open-source models constantly improving, running a quantized open-source LLM on-device will soon be a viable way to improve inference time greatly.

Moreover, while we incorporate two measures specifically designed to improve safety and reliability, deploying autonomous or semi-autonomous RSAs in real-world scenarios still has the potential to bring risks from unexpected circumstances an AI system might not be able to handle.

\section{Conclusion}

We present \framework, a modular framework for natural surgeon-robot interaction. We show that our training-free approach, which uses pre-trained LLMs to provide low-level control of surgical robots, can successfully interact with small surgical objects and execute surgeon commands for automating surgical sub-tasks. Safety is bolstered through re-planning capabilities and a human-in-the-loop approach. We evaluate the efficacy of \framework for common surgical sub-tasks in simulated and physical experiments in the lab and show that the proposed method succeeds across different sub-tasks with various difficulty levels. These results suggest that language-guided autonomy has the potential to enhance surgeon's efficiency in surgical procedures.

In future work, we plan to test the viability of quantized open-source LLMs on-device to improve inference time. This will also address any privacy concerns stemming from transmitting highly sensitive medical information to off-site servers. Furthermore, we intend to explore the usefulness of fine-tuned large language and vision models in \framework.

\bibliographystyle{IEEEtran}
\bibliography{references}

\end{document}